\documentclass[conference]{IEEEtran}
\IEEEoverridecommandlockouts

\usepackage{cite}
\usepackage{amsmath,amssymb,amsfonts}
\usepackage{algorithmic}
\usepackage{graphicx}
\usepackage{textcomp}
\usepackage{xcolor}
\usepackage{tabularx}
\usepackage{stfloats}

\def\BibTeX{{\rm B\kern-.05em{\sc i\kern-.025em b}\kern-.08em
    T\kern-.1667em\lower.7ex\hbox{E}\kern-.125emX}}

\usepackage{url}

\begin{document}

\title{Evaluating the Effectiveness of Persona Simulation in Opinion Prediction with GPT-4.1\\}

\author{\IEEEauthorblockN{Sarah Y. Li}
\IEEEauthorblockA{\textit{McLean High School} \\
McLean, VA, USA \\
sarah.yingqi.li@gmail.com}
\and
\IEEEauthorblockN{Ziyu Yao}
\IEEEauthorblockA{\textit{Department of Computer Science} \\
\textit{George Mason University, VA, USA}\\
ziyuyao@gmu.edu}
}

\maketitle



\begin{abstract}
Persona simulation involves utilizing large language models (LLMs) to anticipate human choices or interactions based on specific characteristic information. To further understand current limitations and future directions, we tested persona simulation in opinion prediction with GPT-4.1 (knowledge cutoff by June 2024). Using personas from nine U.S. states provided by Columbia University's Personas dataset, GPT-4.1 accurately predicted 2024 election outcomes in eight out of the nine states, only failing in one of the swing states. We then focused on opinions related to medicine and healthcare. With the American Trends Panel Wave 123 dataset from Pew Research Center, GPT-4.1 was able to anticipate beliefs about childhood vaccines with an accuracy of up to 0.94. Furthermore, we applied GPT-4.1 to generate conversations among personas and observed that the simulated dialogues and opinions adhered well to personas' personalities and backgrounds, albeit lacking natural human-like flow. Persona simulation proves to be a promising application of artificial intelligence as long as biases are addressed. In the near future, it will be beneficial to apply it to opinion analysis and reaction prediction in diverse fields ranging from public health to lawmaking to economics.
\end{abstract}

\begin{IEEEkeywords}
persona simulation, large language models, generative artificial intelligence, social sciences.
\end{IEEEkeywords}


\section{Introduction}
Persona simulation involves utilizing large language models (LLMs) to anticipate human choices or interactions based on characteristic information about demographics, background, or personality. Generated samples can be used as additional data points in surveys, addressing the challenges that come with data collection, such as imbalanced sampling or non-response bias. Applications include marketing \cite{sarstedt2024marketing}, political science \cite{argyle2023politics}, social science \cite{chang2025networks}, and more. Promise has also been shown by Park et al. in simulating human conversations and interactions in a virtual town environment \cite{park2023generative} and Yue et al. in an educational setting \cite{yue2024mathvc}. This creates possibilities for improved dialogue personalization with chatbots in a variety of use cases.

However, persona simulation has been noted to show bias, overestimate homophily, and assume monolithic qualities for groups \cite{li2025columbia, chang2025networks}. Concerns have also been raised about the difficulty of generating personas that can truly represent the population of interest in terms of intersectional demographics; census data, for instance, only reports marginal distributions without showing the joint distributions across different attributes \cite{li2025columbia}. Representing personality traits is also challenging, and current experiments rely on tests such as the Big Five to capture characteristics \cite{li2025columbia, park2024simulations}.

In this work, we seek to understand the current promises and limitations of LLM-based persona simulation by evaluating GPT-4.1~\cite{openai4.1} in opinion prediction: election forecasting, medical opinion prediction, and dialogue simulation. 
For election prediction, we looked at overall outcomes as well as voting distributions for nine states. Medical opinion prediction involved assessing accuracies for four pertinent questions relating to vaccines and healthcare. For dialogue simulation, we generated conversations between three personas to evaluate how closely dialogues adhered to the given personas.

We found that opinion prediction with GPT-4.1 can reach high accuracies and align closely with the ground truth, but the underlying bias is undeniable, especially with election forecasting. GPT-4.1 relied heavily on over-generalizations of demographic groups when predicting votes. With dialogue generation, personas' personalities were not captured well, and extracted opinions relied on over-generalizations as well. Additionally, generated conversations sounded unrealistic and had a manufactured cheerfulness, limiting the model's potential use for simulations using current methods.

Yet persona simulation still opens many possibilities aside from politics, healthcare, and conversations, including but not limited to forecasting commerce choices \cite{mansour2025stock} and predicting effectiveness ratings of advertisements or public service announcements \cite{sheeran2024vaping}. We hope to see persona simulation grow as existing techniques and frameworks are refined.





\section{Related Work}


As LLMs have become more capable, more research has turned to persona simulation as a viable application. Personas can come in various forms that are used in different situations; as outlined in Chen et al.'s work \cite{chen2024from}, persona types include demographic personas, character personas, and individualized personas. In our project, our focus is on demographic personas. Similar simulations have used personas as ``silicon samples'' to create realistic population representations in contexts of marketing \cite{sarstedt2024marketing}, politics \cite{argyle2023politics, li2025columbia}, social network simulation \cite{chang2025networks}, and more.

Previous work has focused on persona simulation in election prediction. Argyle et al. used the American National Election Studies (ANES) dataset to predict the results of the 2012, 2016, and 2020 elections \cite{argyle2023politics}. Li et al., on the other hand, created personas for each state using census data and used those to predict election results for 2016, 2020, and 2024 \cite{li2025columbia}. However, both did not highlight state-by-state voting distributions, which is a key focus in our work. We also noted important features that impacted GPT-4.1's predictions, providing insight into the reasoning behind decisions.

In addition, the existing literature has had a limited focus on anticipating medical or health-related opinions, instead focusing on simulating patient-doctor interactions \cite{kyung2025patientsim}. In this project, one of our approaches focused specifically on opinions related to vaccines and healthcare in the United States; however, our other approaches study predictions related to political opinions and dialogues, which thus provide a more comprehensive understanding of GPT-4.1's capability for persona simulation.

Dialogue generation has been a highly tested field with the emergence and popularization of LLMs. Chatbots designed to behave like famous celebrities or fictional characters are a common example \cite{CharacterAI}. When communicating with LLMs impersonating personas across ages and expertises, Salewski et al. have noted that impersonation is effective for making conversations more realistic, but may introduce race and gender bias \cite{salewski2023llms}. In this project, we evaluate the ability of GPT-4.1 to represent the personality and beliefs of a persona in the context of a conversation, and evaluate whether any biases exist.


\begin{figure*} [t!]
    \centering
    \includegraphics[width=\linewidth]{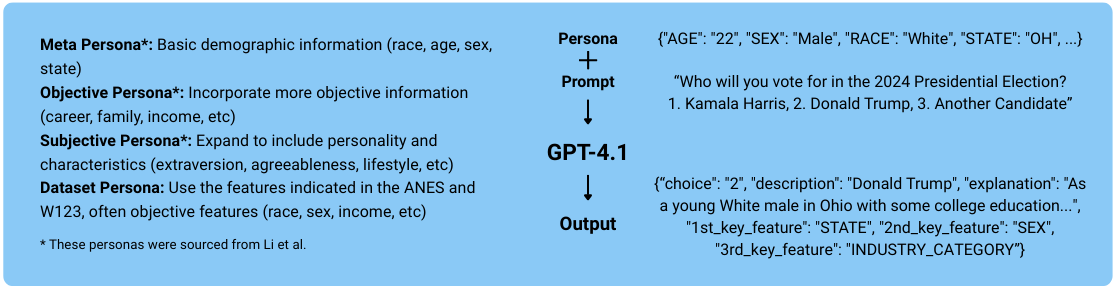}
    \caption{The persona simulation framework.} 
    \label{fig1}
\end{figure*}

\section{Persona Simulation Framework}

Effective persona simulation relies on representing the population using diverse characteristics. The personas we used were extracted from the Personas dataset (PERSONAS) from Columbia University \cite{columbia2025dataset}, the 2024 American National Election Studies (ANES) dataset \cite{anes2024dataset}, and the American Trends Panel Wave 123 (W123) dataset from Pew Research Center \cite{pew2023dataset}.

Overall, we used four types of personas: meta, objective, subjective, and dataset. The first three descriptions below are built upon Li et al.'s definitions, methods, and the PERSONAS dataset they provided \cite{li2025columbia, columbia2025dataset}.
\\

\textbf{Meta Personas:} These provide characteristics that can be found in census data — race, age, and sex — and were developed using realistic census distributions from the 48 states of the United States mainland. Each state has 1000 meta personas.

\textbf{Objective Personas:} These extend meta personas to include more information about demographics and lifestyle. They include characteristics such as income, marital status, education level, household language, and more.

\textbf{Subjective Personas:} In addition to the information from objective personas, these add information about personality, such as Big Five scores, political views, religion, and ability to speak English.

\textbf{Dataset Personas:} These personas use the features provided by a given dataset; in this study, the additional datasets we used were ANES and W123, which had 5521 and 10,701 randomly sampled individuals, respectively. ANES included 2024 election information and W123 included opinions on healthcare.
Most of the information provided by those datasets would be in the realm of objective personas, with features such as income, education level, gender, race, and more.
\\

The additional information in objective and subjective personas was generated by Llama-3.1-70B \cite{MetaLlama3_1_70B} given existing meta personas. For the objective personas, the prompt gave multiple-choice options for each feature. For subjective personas, however, most features were open-ended, with instructions to remain ``reasonable and succinct''. We refer readers to Li et al. \cite{li2025columbia} for more details.

For opinion simulations, personas were fed one by one in their existing json format to GPT-4.1 using the same prompt for all. All questions were treated as forced multiple-choice to prevent GPT from declining to answer. Outputs consisted of the predicted choice, an explanation, and the extracted top three features most contributing to the selected choice. This framework can be seen in Fig.~\ref{fig1}.


\section{Experimental Setup}

\subsection{Election Forecasting}

Election forecasting was conducted using two different approaches: one using personas from PERSONAS \cite{columbia2025dataset}, and the other using personas from ANES \cite{anes2024dataset}.

1000 meta personas and 1000 objective personas from nine states — three red states (Alabama, Texas, Wyoming), three blue states (Virginia, Maryland, California), and three swing states (Pennsylvania, North Carolina, Nevada) — were extracted from PERSONAS. As PERSONAS lacked a ground truth vote for each persona, we relied on state results and voting distributions \cite{electionresults} as evaluation markers.

ANES provided 2024 voting preferences and related information for 5521 adult United States citizens. We extracted 11 columns: sex, gender, race, race of spouse, education level, household income, place of birth, sexual orientation, transgender status, number of children, and vote. We then removed any samples that had missing values for columns, ending up with 2882 people. Of the 2882 people remaining, 1532 voted for Kamala Harris, 1289 for Donald Trump, 15 for Jill Stein, 9 for Cornel West, and 37 for alternate candidates. Using these samples, we then performed persona simulation using multinomial logistic regression and GPT-4.1.

For feature importance analysis with logistic regression, we used absolute values of coefficients to sort importance. For GPT-4.1, as part of the output of each generated prediction, we asked for the vote, an explanation, and the top three key features contributing to the final voting decision.

GPT-4.1 has a knowledge cutoff of June 2024 \cite{openai4.1}, making it suitable for the prediction of the 2024 election. Despite some literature suggesting that using GPT to predict earlier elections does not impact accuracy or reliability, it may still introduce unseen effects \cite{argyle2023politics, li2025columbia}. Thus, we decided to focus only on the 2024 election and ignore elections that occurred earlier.

\subsection{Healthcare Opinion Prediction} 

For opinion prediction, we used W123 \cite{pew2023dataset}, which included various questions about government, healthcare, and emerging technologies. The original dataset included 10,701 patients and 145 questions. After removing questions where greater than 50\% of patients declined to answer and then removing any patients who declined to answer questions, we were left with 7992 patients and 110 questions. We isolated four major questions that we wanted to focus on, pertaining to vaccines and the healthcare system. See Table~\ref{tab:W123} for the questions.

\begin{table}[htbp]
\centering
\caption{Four highlighted questions from W123 to test \\ healthcare opinion simulation}
\begin{tabularx}{0.48\textwidth}{|p{0.06\textwidth}||p{0.09\textwidth}|X|p{0.041\textwidth}|}
 \hline
 Label & Question & Options & \% of 1 \\
 \hline
 CHILD & Overall, do you think... & 
  1: ``The benefits of childhood vaccines for measles, mumps, and rubella outweigh the risks''; 
  2: ``The risks outweigh the benefits'' & 90\% \\
 \hline
 COVID & Overall, do you think... &
 1: ``The benefits of COVID-19 vaccines outweigh the risks''; 2: ``The risks outweigh the benefits'' & 68\%
 \\
 \hline
 CHOICE & Which comes closest to your point of view?
 & 1: ``Parents should be able to decide NOT to vaccinate their children''; 2: ``Healthy children should be required to be vaccinated in order to attend public schools'' & 30\%
 \\
 \hline
 HEALTH & Which comes closest to your point of view?
 & 1: ``Medical treatments these days are worth the costs because they allow people to live longer and better quality lives''; 2: ``Medical treatments these days often create as many problems as they solve'' & 51\%
 \\
 \hline
\end{tabularx}
\label{tab:W123}
\end{table}

For the features of the dataset personas, we extracted 33 objective features for each persona. We then randomly selected 100 persona samples to perform the simulation. The ``\% of 1'' column refers to the percentage of individuals who responded with a choice of ``1'' to the question. For evaluation, in addition to accuracy, we used F1 score to account for class imbalance as seen in some questions in Table~\ref{tab:W123}.

When scores remained low, we incorporated an additional 30 features for each persona, making a total of 63 features. These features provided more information about personas' healthcare habits and beliefs, improving GPT-4.1's performance.

\subsection{Persona Conversation Simulation} 

For conversation generation, we randomly selected 3 subjective personas from PERSONAS and prompted GPT-4.1 to generate a realistic conversation given a topic. The topics we used were education in schools, vaccines, and the treatment of transgender people, as well as more common and unstructured topics: daily routines, hobbies, and ``no topic.''

Knowing that we did not specifically prompt GPT-4.1 to sound human, our evaluations mostly focused on the following two questions:

\begin{enumerate}
    \item Is this conversation realistic? Would humans interact in the way that these personas do?
    \item Are the personas' personalities and backgrounds reflected in their simulated words and responses to others?
\end{enumerate}


\section{Experimental Results}

\subsection{Election Forecasting}

\begin{figure} [t!]
    \centering
    \includegraphics[width=\linewidth]{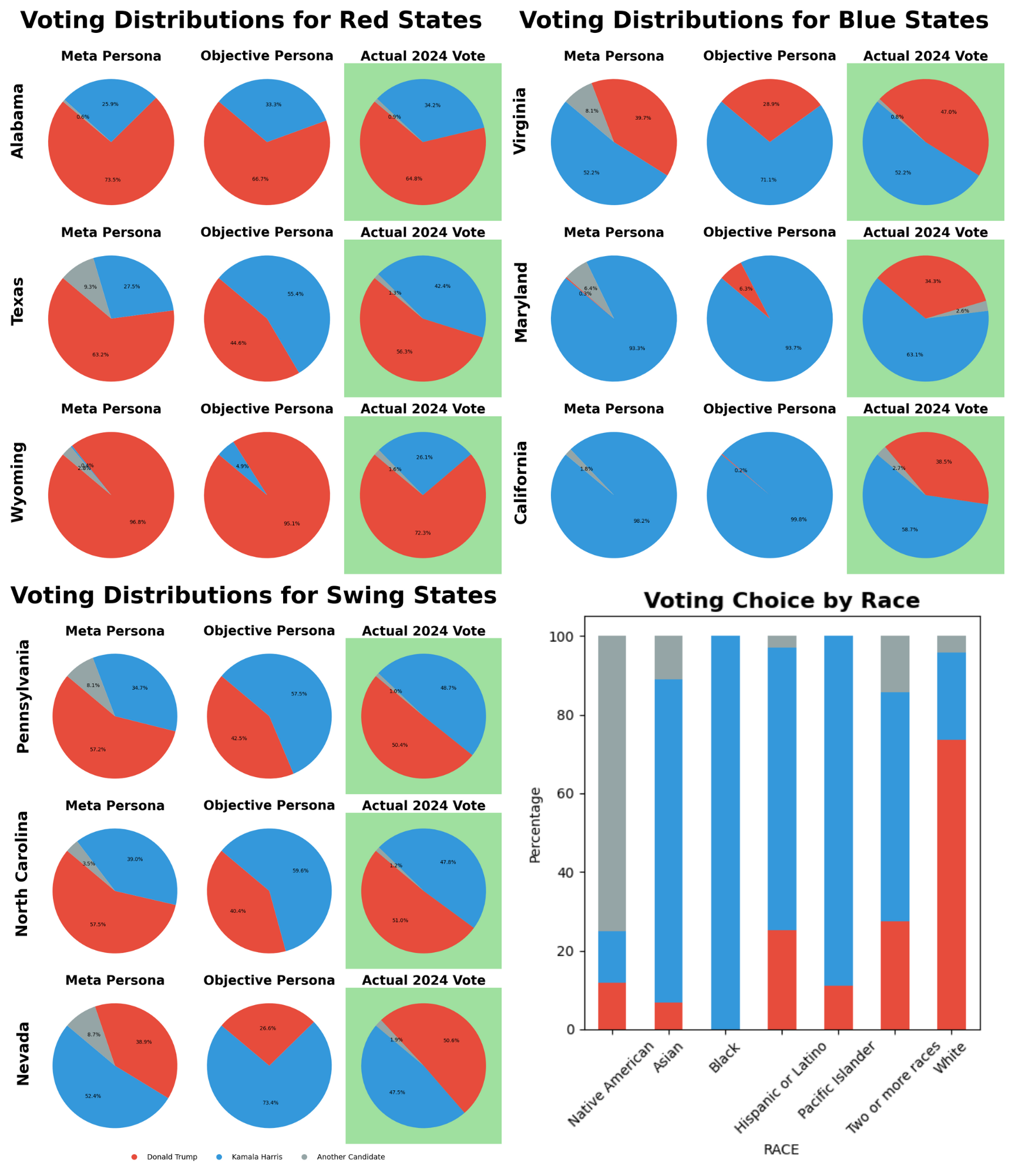}
    \caption{Voting distributions for nine states and across races.}
    \label{fig2}
\end{figure}

With samples from PERSONAS, we used GPT-4.1 to predict votes for nine different states in the 2024 election (Trump vs. Harris), the distributions of which are shown in Fig.~\ref{fig2}. Using meta personas, eight of the nine state outcomes were predicted correctly, with the exception of Nevada, a swing state. Using objective personas, however, only five out of nine state outcomes were predicted correctly. In all states, the predicted objective persona state distribution had a higher percentage of blue votes. A similar phenomenon was observed by Li et al., so this may be due to bias in the generation of objective personas or bias in the voting simulation \cite{li2025columbia}. This may suggest that including too much information in personas may lead to misleading results, and balance is needed.

Despite successfully predicting voting outcomes in eight states using meta personas, voting distributions differ from the ground truth. In states that have historically voted red or blue, the difference was especially apparent. In California, for example, 98.2\% of personas were predicted to vote blue, with no red votes. In the actual 2024 election, 58.7\% of votes went to Harris, and 38.5\% voted for Trump. A similar trend occurs with Maryland, another blue state, and Wyoming, a red state. In general, predictions also overestimate the proportion of votes going to third-party candidates.

Digging deeper into the distribution of votes by race using meta personas, more trends become clear. All Black voters were predicted to support Harris, in comparison to the ground truth of 83\% \cite{pew2025election}. Additionally, 75\% of Native Americans were predicted to vote for third parties, and 73.5\% of Whites for Trump, both of which are overestimates. These show the biases that impact the true performance of persona simulation, underneath the state-by-state results.

\begin{figure} [t!]
    \centering
    \includegraphics[width=\linewidth]{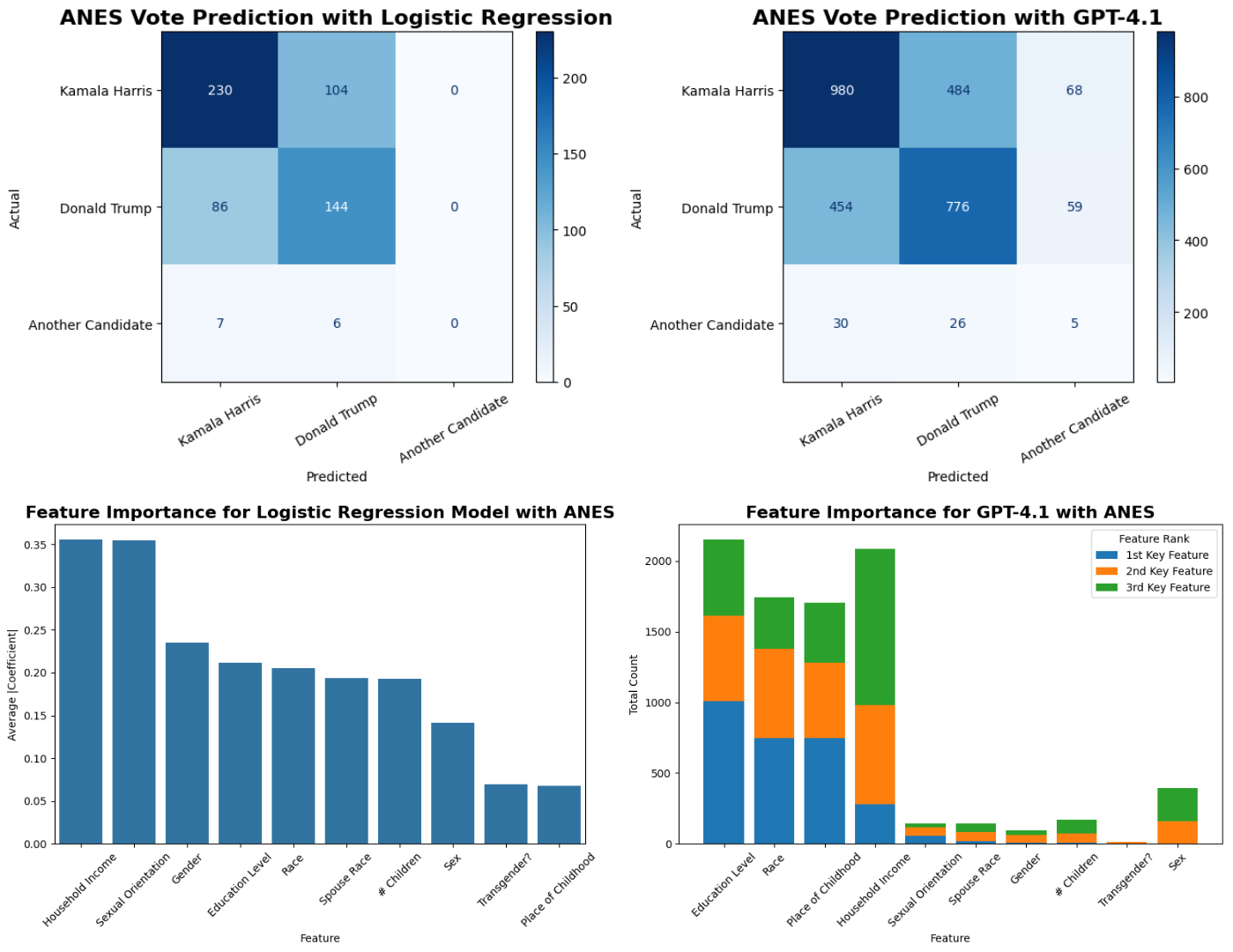}
    \caption{Confusion matrices and feature importances for ANES using logistic regression and GPT-4.1.}
    \label{fig3}
\end{figure}

Persona simulation with ANES followed a different format, as voting truth labels were given for each sample. Our main goal was to test logistic regression and GPT-4.1 as viable options for persona simulation and also analyze the important features used by each model. Overall, logistic regression had an accuracy of 0.648, and GPT-4.1 had an accuracy of 0.610, showing room for improvement. Confusion matrices and important features can be seen in Fig.~\ref{fig3}.

The four most important features for logistic regression were household income, sexual orientation, gender, and education level. On the other hand, the top four features for GPT-4.1 were education level, household income, race, and place of childhood (referring to where the persona grew up). For logistic regression, the features mostly follow a smooth decline of importance, but for GPT-4.1, the top four features are considered very heavily compared to the rest. Since GPT-4.1 performs worse than logistic regression, this may indicate that GPT is focusing on the wrong features, causing it to struggle. These reveal hidden influences underlying every prediction, and emphasize the importance of bias analysis for all models to increase transparency.

\subsection{Healthcare Opinion Prediction}

When using GPT-4.1 to predict healthcare-related opinions from W123, scores were difficult to improve, especially for questions with split responses, such as HEALTH (refer to Table~\ref{tab:W123}). After adding more information about each persona's vaccination habits and other healthcare-related beliefs, scores skyrocketed, as shown in Table~\ref{tab:W123res}. At its peak, GPT-4.1 was able to anticipate beliefs about childhood vaccines with an accuracy of up to 0.94 and an F1 score of up to 0.8468.

\begin{table}[b!]
\centering
\caption{The accuracies and F1 scores of our W123 questions before and after adding additional healthcare information}
\begin{tabular}{|l|c|c|c|c|}
 \hline
  & \multicolumn{2}{c|}{Before} & \multicolumn{2}{c|}{After} \\
 \hline
  Label & Acc & F1 & Acc & F1 \\
 \hline
 CHILD  & 0.89 & 0.6036 & 0.94 & 0.8468 \\
 COVID  & 0.82 & 0.7532 & 0.88 & 0.8698 \\
 CHOICE & 0.75 & 0.6710 & 0.81 & 0.7759 \\
 HEALTH & 0.59 & 0.5890 & 0.75 & 0.7442 \\
 \hline
\end{tabular}
\label{tab:W123res}
\end{table}

Across Table~\ref{tab:W123res}, it is clear that the HEALTH question (referring to whether or not medical treatments are worth the costs or not) causes GPT-4.1 to perform the worst. It is a very divisive question, with 51\% believing treatments are worth the costs, while 49\% disagree. It is also affected by personal experiences, which are difficult to capture through personas. Thus, we realize that GPT-4.1 struggles most with controversial questions where most people are evenly split.

\subsection{Persona Conversation Simulation}
 
Dialogue generation involved randomly selecting three subjective personas from PERSONAS and prompting them to have a conversation around a provided topic. We used a combination of structured and unstructured topics, ranging from the education system to hobbies.

In general, we noticed that the simulated conversations were very rigid, with each persona taking turns to speak without variation in the order of speaking. For one particular topic, which was about the treatment of transgender people, the conversation read like a scripted refute of transphobia, with one persona raising concerns and the other two swiftly batting them down. Other similar dialogues contributed to decreasing the realism of generated conversations.

The beliefs that the conversations showed adhered to the personas, and each persona showed unique hobbies and occupations. GPT-4.1 was acceptable at portraying objective traits, but struggled with more subjective ones.
No matter what personality each persona had, the dialogue read the same. Rarely was there an emotion other than optimism, and introverted and extroverted personas showed no difference in their manner of speaking.

\section{Conclusion and Future Work}

Persona simulation can reach high accuracies, but a closer look into those results reveals bias and overgeneralization. For election forecasting, we observed biases by state and race, and GPT-4.1 focused on different persona features than the more impartial logistic regression. Additionally, current models have room for improvement in generating conversations that sound more realistic and adhere to given personalities.

Park et al. has an effective framework in which individuals are interviewed on their personal experiences, which are then used to predict responses to questions \cite{park2024simulations}. That approach eliminates bias that may occur when LLMs generate subjective traits for personas; indeed, Li et al. noted that as LLMs had more influence in the generation of personas, those personas tended to lean left, have skewed opinions, and include more positivity \cite{li2025columbia}. Park et al.'s approach also incorporates more life experiences and personality compared to more tabular approaches for capturing information about a persona \cite{park2024simulations}. We would like to perform future work with that framework, given more time and resources.

Developing sophisticated metrics to assess generated dialogues is also important, as our current methods remain quite subjective. More objective analyses would bring credence to our findings and allow for scientific growth.

In the future, more work needs to be done in bias analysis, especially looking at different opinion distributions for each demographic group. Increasing accuracy and other evaluation metrics is important and exciting, but should not come at the cost of transparency or fairness.

Overall, persona simulation is a promising use of artificial intelligence as long as biases are addressed. It will be beneficial to apply it to opinion and reaction prediction in diverse fields, from lawmaking to economics to quality assurance, enabling more informed decisions and personalized interventions.


\section*{Acknowledgment}
We gratefully acknowledge the administrative and financial support of George Mason University's Aspiring Scientists Summer Internship Program (ASSIP), which facilitates hosting high school students in Mason research groups.

\bibliographystyle{IEEEtran}
\bibliography{custom}

@article{chen2024from,
title={{From Persona to Personalization: A Survey on Role-Playing Language Agents}},
author={Jiangjie Chen and Xintao Wang and Rui Xu and Siyu Yuan and Yikai Zhang and Wei Shi and Jian Xie and Shuang Li and Ruihan Yang and Tinghui Zhu and Aili Chen and Nianqi Li and Lida Chen and Caiyu Hu and Siye Wu and Scott Ren and Ziquan Fu and Yanghua Xiao},
journal={Transactions on Machine Learning Research},
issn={2835-8856},
year={2024},
}

@article{yue2024mathvc,
  title={{MathVC: An LLM-Simulated Multi-Character Virtual Classroom for Mathematics Education}},
  author={Yue, Murong and Lyu, Wenhan and Mifdal, Wijdane and Suh, Jennifer and Zhang, Yixuan and Yao, Ziyu},
  journal={arXiv preprint arXiv:2404.06711},
  year={2024}
}

@inproceedings{park2023generative,
  title={{Generative agents: Interactive simulacra of human behavior}},
  author={Park, Joon Sung and O'Brien, Joseph and Cai, Carrie Jun and Morris, Meredith Ringel and Liang, Percy and Bernstein, Michael S},
  booktitle={{Proceedings of the 36th annual acm symposium on user interface software and technology}},
  pages={{1--22}},
  year={2023}
}

@inproceedings{sarstedt2024marketing,
    author = {Sarstedt, Marko and Adler, Susanne and Rau, Lea and Schmitt, Bernd},
    title = {{Using large language models to generate silicon samples in consumer and marketing research: Challenges, opportunities, and guidelines}},
    booktitle = {{Psychology \& Marketing}},
    pages = {{1254--1270}},
    year = {{2024}}
}

@inproceedings{chang2025networks,
    author = {Chang, Serena and Chaszczewicz, Alicja and Wang, Emma and Josifovska, Maya and Pierson, Emma and Leskovec, Jure},
    title = {{LLMs Generate Structurally Realistic Social Networksbut Overestimate Political Homophily}},
    booktitle = {{Proceedings of the Nineteenth International AAAI Conference on Web and Social Media}},
    pages = {{341--371}},
    year = {{2025}}
}

@inproceedings{li2025columbia,
    author = {Li, Ang and Chen, Haozhe and Namkoong, Hongseok and Peng, Tianyi},
    title = {{LLM Generated Persona is a Promise with a Catch}},
    booktitle = {{arXiv preprint arXiv:2503.16527}},
    year = {{2025}}
}

@inproceedings{argyle2023politics,
    author = {Argyle, Lisa P and Busby, Ethan C and Fulda, Nancy and Gubler, Joshua and Rytting, Christopher and Wingate, David},
    title = {{Out of One, Many: Using Language Models to Simulate Human Samples}},
    booktitle = {{Political Analysis}},
    pages = {{337--351}},
    year = {{2023}}
}

@inproceedings{park2024simulations,
    author = {Park, Joon Sung and Zou, Carolyn Q and Shaw, Aaron and Hill, Benjamin Mako and Cai, Carrie and Morris, Meredith Ringel and Willer, Robb and Liang, Percy and Bernstein, Michael S},
    title = {{Generative Agent Simulations of 1,000 People}},
    booktitle = {{arXiv preprint arXiv:2411.10109}},
    year = {{2024}}
}

@inproceedings{salewski2023llms,
    author = {Salewski, Leonard and Alaniz, Stephan and Rio-Torto, Isabel and Schulz, Eric and Akata, Zeynep},
    title = {{In-Context Impersonation Reveals Large Language
Models’ Strengths and Biases}},
    booktitle = {{Proceedings of the 37th International Conference on Neural Information Processing Systems}},
    pages = {{72044 -- 72057}},
    year = {{2023}}
}

@electronic{anes2024dataset,
    author = {{{American National Election Studies}}},
    title = {{ANES 2024 Time Series Study Full Release [dataset and documentation]}},
    url = {https://electionstudies.org/data-center/2024-time-series-study/},
    year = {{2025}}
}

@electronic{pew2023dataset,
    author = {{{Pew Research Center}}},
    title = {{American Trends Panel Wave 123}},
    url = {https://www.pewresearch.org/dataset/american-trends-panel-wave-123/},
    year = {{2023}}
}

@electronic{columbia2025dataset,
    author = {{Tianyi-Lab}},
    title = {{Personas}},
    url = {https://huggingface.co/datasets/Tianyi-Lab/Personas},
    year = {2025}
}

@inproceedings{pew2025election,
    author = {Hartig, Hannah and Keeter, Scott and Daniller, Andrew and Van Green, Ted},
    title = {{Voting patterns in the 2024 election}},
    booktitle = {Pew Research Center},
    year = {2025}
}

@electronic{openai4.1,
    author = {OpenAI},
    title = {{Introducing GPT-4.1 in the API}},
    year = {2025},
    url = {https://openai.com/index/gpt-4-1/}
}

@electronic{electionresults,
    author = {{CNN Politics}},
    title = {{Election 2024: Presidential results}},
    year = {2024},
    url = {https://www.cnn.com/election/2024/results/president?election-data}
}

@misc{MetaLlama3_1_70B,
  author       = {{Meta AI}},
  title        = {Llama-3.1-70B Large Language Model},
  howpublished = {\url{https://huggingface.co/meta-llama/Llama-3.1-70B}},
  year         = {2024},
  note         = {Released July 23, 2024},
}

@misc{CharacterAI,
  author       = {Noam Shazeer and Daniel de Freitas},
  title        = {{Character.AI}},
  howpublished = {\url{https://character.ai/}},
  year         = {2025},
}

@inproceedings{mansour2025stock,
    author = {Saab Mansour and Leonardo Perelli and Lorenzo Mainetti and George Davidson and Stefano D'Amato},
    title = {{PAARS: Persona Aligned Agentic Retail Shoppers}},
    booktitle = {{Proceedings of the 1st Workshop for Research on Agent Language Models (REALM 2025)}},
    year = {2025},
    pages = {143--159}
}

@inproceedings{sheeran2024vaping,
    author = {Paschal Sheeran and Alexander Kenny and Andrea Bermudez and Kurt Gray and Emily F Galper and Marcella Boynton and Seth M Noar},
    title = {{Artificial Intelligence Simulation of Adolescents’ Responses to Vaping-Prevention Messages}},
    booktitle = {{JAMA Pediatrics}},
    year = {2024}
}

@inproceedings{kyung2025patientsim,
    author = {Daeun Kyung and Hyunseung Chung and Seongsu Bae and Jiho Kim and Jae Ho Sohn and Taerim Kim and Soo Kyung Kim and Edward Choi},
    title = {{PATIENTSIM: A Persona-Driven Simulator for Realistic Doctor-Patient Interactions}},
    booktitle = {arXiv preprint arXiv:2505.17818},
    year = {2025}
}

\end{document}